\pdfoutput=1
\documentclass[9pt,shortpaper,twoside,web]{ieeecolor}
\usepackage{etoolbox}

\makeatletter
\@ifundefined{color@begingroup}%
{\let\color@begingroup\relax
\let\color@endgroup\relax}{}%
\def\fix@ieeecolor@hbox#1{%
\hbox{\color@begingroup#1\color@endgroup}}
\patchcmd\@makecaption{\hbox}{\fix@ieeecolor@hbox}{}{\FAILED}
\patchcmd\@makecaption{\hbox}{\fix@ieeecolor@hbox}{}{\FAILED}

\usepackage{generic}
\usepackage{cite}
\usepackage{amsmath,amssymb,amsfonts}
\usepackage{algorithmic}
\usepackage{graphicx}
\usepackage{textcomp}
\usepackage{booktabs}
\usepackage{multirow}
\usepackage{verbatim}
\usepackage{stfloats}
\usepackage{array}
\usepackage{algorithm}
\usepackage{url}

\def\BibTeX{{\rm B\kern-.05em{\sc i\kern-.025em b}\kern-.08em
    T\kern-.1667em\lower.7ex\hbox{E}\kern-.125emX}}
\begin{document}

\title{Human Locomotion Implicit Modeling Based Real-Time Gait Phase Estimation}
\author{Yuanlong Ji, Xingbang Yang, Ruoqi Zhao, Qihan Ye, Quan Zheng, and Yubo Fan
\thanks{Manuscript received April 30, 2025. Research supported by Beijing Natural Science Foundation (Grant number L222139), the National Natural Science Foundation of China ((Grant number 52475291) and the Fundamental Research Funds for the Central Universities (Grant numbers YWF-23-Q-1031 and JKF-YG-22-B010).\textit{(Yuanlong Ji and Xingbang Yang are co-first authors.)(Corresponding authors: Xingbang Yang; Yubo Fan.)}}
\thanks{This work involved human subjects or animals in its research. The approval of all ethical and experimental procedures and protocols was granted by the Ethics Committee of Beihang University under No. BM20220181 on August 28, 2023, and was performed in accordance with the Declaration of Helsinki.}
\thanks{The authors are with the Key Laboratory of Biomechanics and Mechanobiology (Beihang University), Ministry of Education; Key Laboratory of Innovation and Transformation of Advanced Medical Devices, Ministry of Industry and Information Technology; National Medical Innovation Platform for Industry-Education Integration in Advanced Medical Devices (Interdiscipline of Medicine and Engineering); School of Biological Science and Medical Engineering, Beihang University, Beijing, 100191, China (jiyuanlong@buaa.edu.cn; yangxingbang@buaa.edu.cn; yubofan@buaa.edu.cn)}
}

\maketitle

\begin{abstract}
Gait phase estimation based on inertial measurement unit (IMU) signals facilitates precise adaptation of exoskeletons to individual gait variations. However, challenges remain in achieving high accuracy and robustness, particularly during periods of terrain changes. To address this, we develop a gait phase estimation neural network based on implicit modeling of human locomotion, which combines temporal convolution for feature extraction with transformer layers for multi-channel information fusion. A channel-wise masked reconstruction pre-training strategy is proposed, which first treats gait phase state vectors and IMU signals as joint observations of human locomotion, thus enhancing model generalization. Experimental results demonstrate that the proposed method outperforms existing baseline approaches, achieving a gait phase RMSE of 2.729 ± 1.071\% and phase rate MAE of 0.037 ± 0.016\% under stable terrain conditions with a look-back window of 2 seconds, and a phase RMSE of 3.215 ± 1.303\% and rate MAE of 0.050 ± 0.023\% under terrain transitions. Hardware validation on a hip exoskeleton further confirms that the algorithm can reliably identify gait cycles and key events, adapting to various continuous motion scenarios. This research paves the way for more intelligent and adaptive exoskeleton systems, enabling safer and more efficient human-robot interaction across diverse real-world environments.
\end{abstract}


\section{Introduction}
\label{sec:introduction}
Gait phase detection based on inertial measurement unit (IMU) signals has been widely applied in controllers for powered exoskeletons~\cite{ref1,ref2,ref3,ref4} and prostheses~\cite{ref5,ref6,ref7} due to its advantages of low cost, portability, and freedom from spatial constraints. However, since the duration of each step in human locomotion naturally varies, directly parameterizing joint angles or torques as functions of time remains highly challenging~\cite{ref8,ref9}. Considering the inherent periodicity of human gait during continuous walking, a common strategy is to decouple spatial and temporal components by modeling joint angles or torques as functions of gait phase instead of time~\cite{ref10,ref11}.

Existing methods for determining gait phase can be broadly categorized into gait phase recognition (GPR) and gait phase estimation (GPE). GPR divides a complete gait cycle into several discrete stages, such as heel strike (HS), toe strike (TS), heel off (HO), toe off (TO), and mid-swing~\cite{ref12}. In contrast, GPE aims to provide a continuous and fine-grained representation of the gait phase rather than a sequence of discrete labels~\cite{ref13}. This continuous representation enables real-time feedback on the current position within the gait cycle, which can help exoskeletons adapt more precisely to individual motion patterns and gait variability, and in theory, achieve better assistive performance than GPR~\cite{ref14}.

Robust gait phase estimation algorithms commonly fall into three categories: (1) \textit{Time-Based Estimation (TBE)}, which estimates gait phase as a proportion of heel contact and stride duration~\cite{ref15}. While this method performs reliably on treadmills at constant speeds, its accuracy significantly degrades when users change their walking speed or gait mode. (2) \textit{Adaptive Oscillator (AO)-based methods}, such as Ronsse \textit{et al.}, who developed an adaptive oscillator to extract high-level features from periodic inputs using only encoder signals from the joint exoskeleton for gait phase estimation~\cite{ref16}; and Zheng \textit{et al.}, who combined capacitive sensing with AO, reporting a root mean square error (RMSE) of 0.19 rad (3.0\% of the gait cycle) under constant-speed walking and 0.31 rad (4.9\%) under variable-speed conditions in experiments with 12 participants~\cite{ref17}. (3) \textit{Data-driven models and neural networks}, such as Medrano \textit{et al.}, who proposed a data-driven kinematic modeling approach using IMU signals to estimate gait phase, phase rate, stride length, and ground slope in real time, achieving a gait phase RMSE of 4.8~$\pm$~2.4\% and a phase rate error of 2.1~$\pm$~0.5\%, with consistent performance on uneven terrain~\cite{ref18}. Kang \textit{et al.} proposed a convolutional neural network (CNN)-based method, achieving a gait phase RMSE of 4.37~$\pm$~0.68\%, improving accuracy by 62.62~$\pm$~5.62\% (\(p{<}0.05\)) compared to TBE~\cite{ref19}. Hong \textit{et al.} introduced a segmented linear label strategy combined with LSTM to enhance speed adaptability, reducing mid- and high-speed gait phase estimation errors by 37.7\%, 43.4\%, and 35.2\%, and low-speed errors by 27.1\% (though without statistical significance)~\cite{ref20,ref21}.

Although time-based estimation methods perform well during steady-state walking, their error rates can reach up to 20\% in dynamic scenarios such as terrain transitions. With the rapid development of deep learning in recent years, leveraging IMU or joint encoder signals in combination with neural networks for accurate gait phase estimation has become a research focus in the exoskeleton field~\cite{ref22,ref23,ref24,ref25}. However, current methods still face limitations in terms of estimation accuracy, robustness across different movement conditions, and computational efficiency~\cite{ref18,ref27}. Therefore, there is an urgent need to design more accurate, robust, and efficient gait phase estimation methods to enhance exoskeleton assistance in complex environments.

Most existing gait phase estimation methods follow a framework where kinematic signals (IMU or encoder data) are used as inputs to predict the current gait phase as the output. In essence, both gait phase and kinematic signals are representations of the underlying human locomotion. Based on this objective reality, this study is the first to propose treating gait phase not only as a target state but also as an observable quantity. Specifically, gait phase is incorporated as an input during pre-training through a channel-wise masked reconstruction approach. This enables implicit modeling of human locomotion as the relationship between different observations, for which neural networks serve as a powerful modeling tool. Experimental results demonstrate that the proposed algorithm significantly outperforms baseline methods. Finally, hardware validation on a hip exoskeleton confirms that the proposed method can efficiently identify different gait cycles and gait events under both stable and varying terrain conditions.

The remainder of this paper is organized as follows. Section~\uppercase\expandafter{\romannumeral2} presents the proposed gait phase estimation network based on implicit modeling of human locomotion. Section~\uppercase\expandafter{\romannumeral3} describes the experimental design and analyzes the results in detail. Section~\uppercase\expandafter{\romannumeral4} validates the practical application of the proposed method through hardware experiments on a hip exoskeleton. Finally, Section~\uppercase\expandafter{\romannumeral5} concludes the paper and discusses future directions and potential improvements for gait phase estimation methods.

\section{Methods}

\subsection{Sensor Configuration and Dataset}
\begin{figure}[!t]
\centering
\includegraphics[width=3.5in]{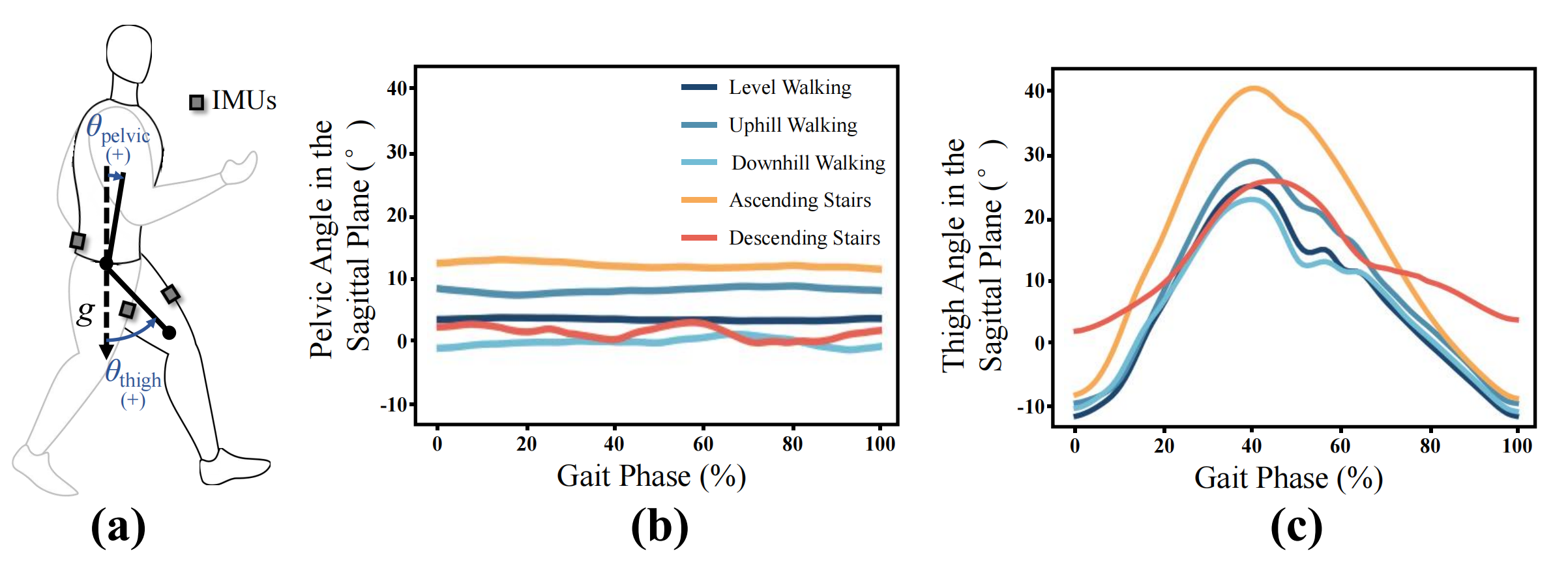}
\caption{Sensor configuration, hip joint angle definitions, and dataset reference curves. (a) IMU layout and the definitions of hip joint, thigh, and pelvis angles in the sagittal plane. (b) Variations of pelvis angle with gait phase. (c) Variations of thigh angle with gait phase.}
\label{fig_1_1}
\end{figure}

This study utilizes publicly available datasets from our previous research~\cite{ref28} for training and validating the proposed gait phase estimation neural network based on implicit modeling of human locomotion. Briefly, the dataset consists of real-world walking data collected from 10 subjects in an out-of-the-lab urban environment, involving five types of terrains: level ground, stair ascent, stair descent, slope ascent, and slope descent. Kinematic data of the hip joint were measured using three IMUs. These three identical IMUs were placed on the front of the left thigh, the front of the right thigh, and the posterior pelvis of the subjects. As shown in Fig.~\ref{fig_1_1}, the hip joint angle in the sagittal plane is defined as the angle between the pelvis and the thigh. When using the gravity direction as a reference, the pelvis angle is defined as the angle between the pelvis and the gravity, and similarly, the thigh angle is defined as the angle between the thigh and the gravity. The anterior rotation of the pelvis and thigh is defined as the positive direction, as illustrated in Fig.~\ref{fig_1_1}(a). The mean variations of the thigh angle and pelvis angle with respect to the gait phase in different terrains are shown in Fig.~\ref{fig_1_1}(b) and Fig.~\ref{fig_1_1}(c), respectively. All procedures involving human participants were reviewed and approved by the Ethics Committee of Beihang University under approval number BM20220181 on August 28, 2023, and were conducted in accordance with the Declaration of Helsinki.

\normalcolor

\subsection{Polar Representation of Gait Phase State}
Using gait phase alone may lead to the loss of temporal information. To address this issue, inspired by the work of Medrano \textit{et al.}~\cite{ref18}, this study employs both the gait phase and its first-order derivative with respect to time to describe the gait phase state. The gait phase state thus consists of two components: the gait phase itself and the gait phase rate. Considering that the gait phase typically varies linearly with time within each step, the gait phase rate at the \( n \)-th sampling point is calculated as follows:
\begin{equation}
\varphi'(n) = \frac{100\%}{L}
\end{equation}
where \( L \) denotes the number of sampling points within the current step. The gait phase at the \( n \)-th sampling point is calculated as follows:
\begin{equation}
\varphi(n) = \frac{L_n}{L} \times 100\%
\end{equation}
where \( L_n \) represents the number of sampling points from the start of the current step to the \( n \)-th sampling point. This definition of gait phase causes a discontinuity at the end of each step, where the gait phase drops from 100\% to 0\%, resulting in a non-differentiable point that hinders effective learning by neural networks. Therefore, inspired by the work of Kang \textit{et al.}~\cite{ref27}, this study adopts a polar coordinate representation of gait phase to ensure continuity, and for the first time integrates the gait phase rate into the representation. The final gait phase state vector \( \boldsymbol{G} \) is expressed as a three-dimensional vector:
\begin{equation}
\boldsymbol{G} = \left[ \cos(2\pi \varphi(n)), \, \sin(2\pi \varphi(n)), \, \varphi'(n) \right]
\end{equation}

\subsection{IMU Time-Series Analysis Network}
\begin{figure*}[!t]
\centering
\includegraphics[width=0.85\textwidth]{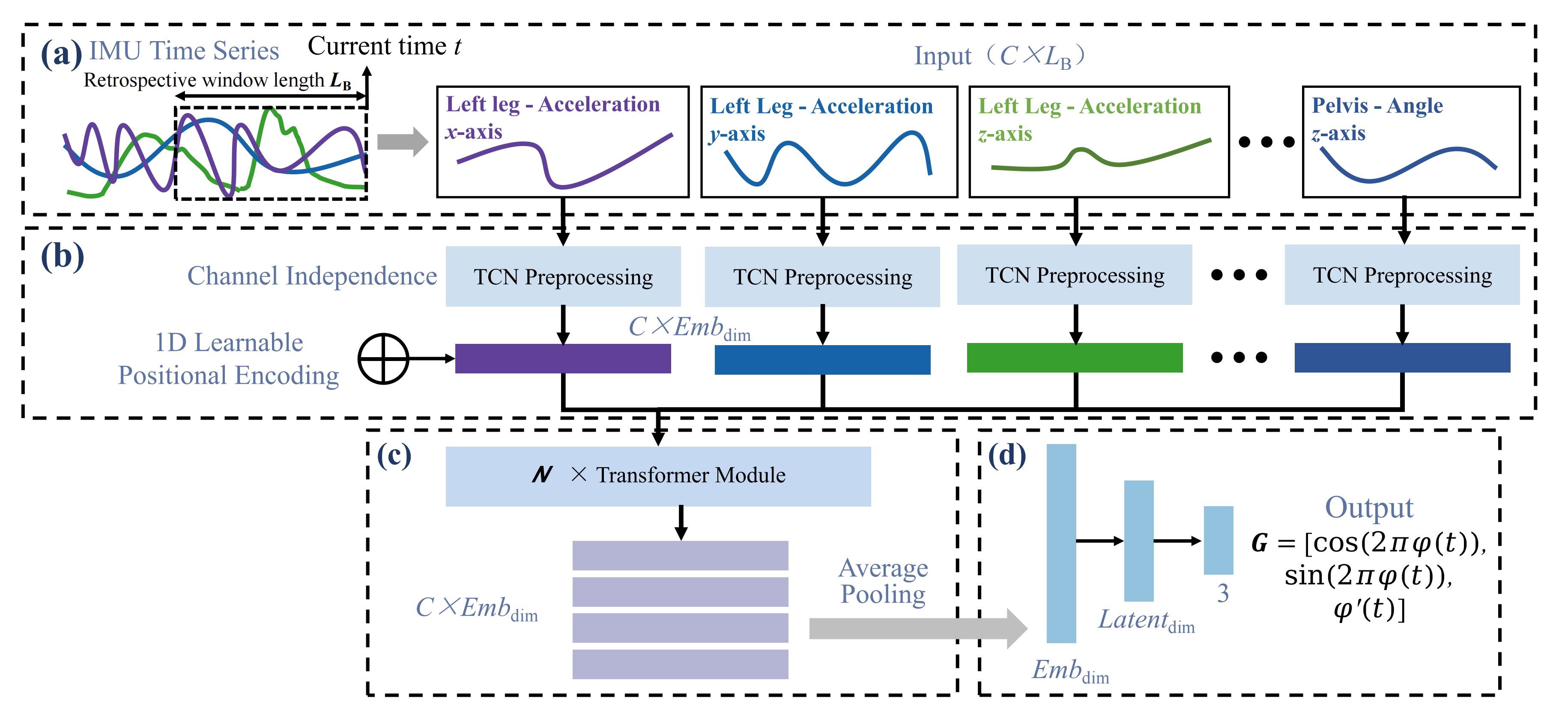}
\caption{Architecture of the gait phase estimation network TCTST. (a) The input consists of multi-channel IMU time-series data, with a length determined by the look-back window \( L_\mathrm{B} \). (b) The input is first processed by the embedding layer. (c) A stack of \( N \) Transformer layers extracts features and handles inter-channel information, followed by average pooling along the channel dimension. (d) A feedforward network maps the extracted features to the desired gait phase state vector \( \boldsymbol{G} \).}
\label{fig_1}
\end{figure*}

This study proposes an IMU time-series analysis network based on Temporal Convolutional Network (TCN)~\cite{ref29} and Transformer~\cite{ref30}, named TCTST (Temporal Convolutional Time Series Transformer). The overall architecture of the proposed network is illustrated in Fig.~\ref{fig_1}. The network consists of three main components: a channel-wise embedding layer, Transformer layers, and a feedforward layer.

\normalcolor

The input to the proposed network consists of multi-channel IMU time-series data. At each sampling point, a fixed-length window of \( L_\mathrm{B} \) data points is extracted using a sliding window approach, where the window moves forward with a stride of 1. Therefore, the network is designed to estimate the gait phase state vector for every sampling point. The IMU time-series data include a total of 21 channels, comprising tri-axial acceleration, tri-axial angular velocity, and pitch angle measurements from IMUs attached to the left leg, right leg, and pelvis. Furthermore, as discussed in Section~\uppercase\expandafter{\romannumeral2}-D, this study also considers incorporating the gait phase state vector as part of the input during pre-training. Thus, the total number of input channels is denoted as \( C \).

\begin{figure}[!t]
\centering
\includegraphics[width=3.5in]{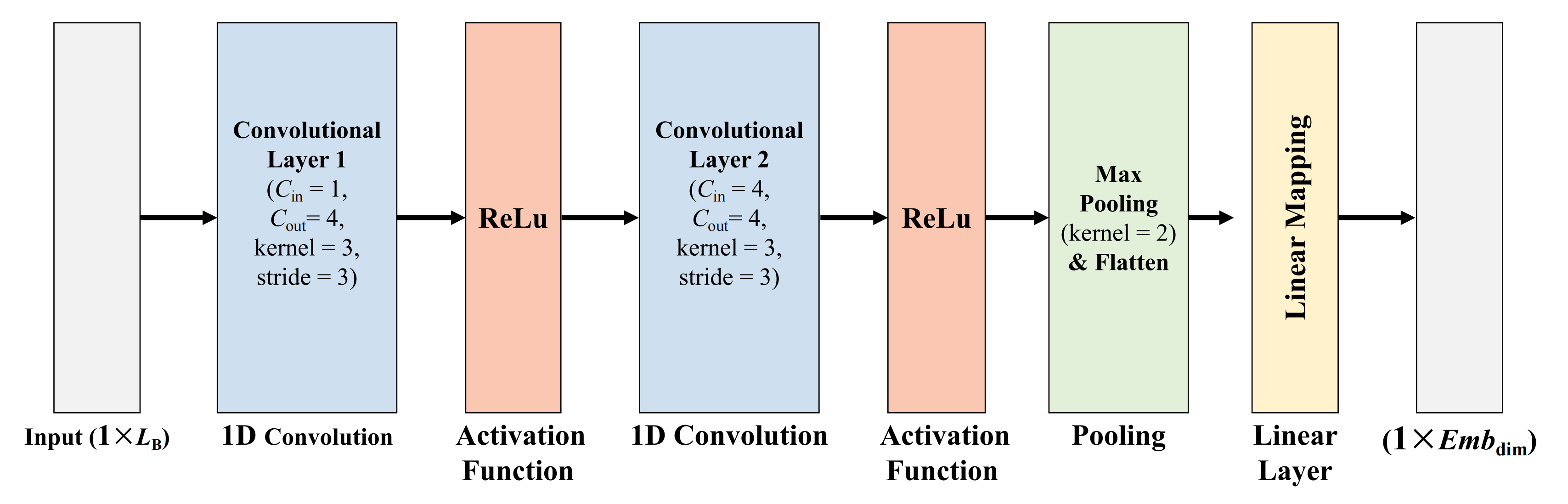}
\caption{Architecture and parameters of the TCN module. Here, \( C_{\mathrm{in}} \) denotes the number of input channels, \( C_{\mathrm{out}} \) denotes the number of output channels, \textit{kernel} represents the kernel size of the convolution or pooling operation, and \textit{stride} indicates the stride of the convolution kernel.}
\label{fig_2}
\end{figure}

The embedding layer consists of two components: a preprocessing module and a positional encoding module. Each channel is first normalized, and then processed independently through a channel-wise TCN, whose detailed network architecture is illustrated in Fig.~\ref{fig_2}. The positional encoding is implemented as a one-dimensional learnable parameter with dimensionality equal to \( C \). This positional encoding is broadcasted along the channel dimension to form a two-dimensional vector of shape \( [C, Emb_\mathrm{dim}] \), which is then added to the TCN-processed feature vector. Here, \( Emb_\mathrm{dim} \) denotes the embedding dimension of the time-series data. After feature fusion through Transformer layers, a two-layer feedforward network maps the features to the output gait phase state vector. The hidden layer of the feedforward network has a dimensionality of \( Latent_\mathrm{dim} \).

In this study, a standard Transformer module design is adopted to construct the Transformer layers for processing IMU time-series data. Each Transformer layer mainly consists of a Multi-Head Attention mechanism and a Feed-Forward Network (FFN). The implementation of the Transformer layers follows default parameter settings, including \textit{MLP\_ratio}, activation function, and \textit{qkv\_bias}. Specifically, \textit{MLP\_ratio} controls the dimensionality of the hidden layer in the FFN, and is set to 4.0, meaning that the hidden layer has a dimensionality four times that of the input. This configuration achieves a good trade-off between model capacity and computational cost. The ReLU activation function is applied to introduce non-linear transformations, thereby enhancing the model’s ability to learn complex feature patterns. The \textit{qkv\_bias} parameter determines whether to include bias terms when computing queries (Q), keys (K), and values (V); in this study, bias terms are included to improve the model's flexibility. 
To optimize the performance of the Transformer layers in gait phase estimation, the number of attention heads \textit{n\_heads} and the number of Transformer layers \( N \) are treated as tunable hyperparameters. Increasing \textit{n\_heads} and \( N \) can enhance the model’s representational capacity, but also requires more training data to prevent underfitting. Therefore, a hyperparameter search experiment is conducted to determine the optimal model size. The implementation details and results can be found in Section~\uppercase\expandafter{\romannumeral3}-C.

\normalcolor

\subsection{Implicit Modeling of Human Locomotion via Mask Modeling}
\begin{figure*}[!t]
\centering
\includegraphics[width=0.85\textwidth]{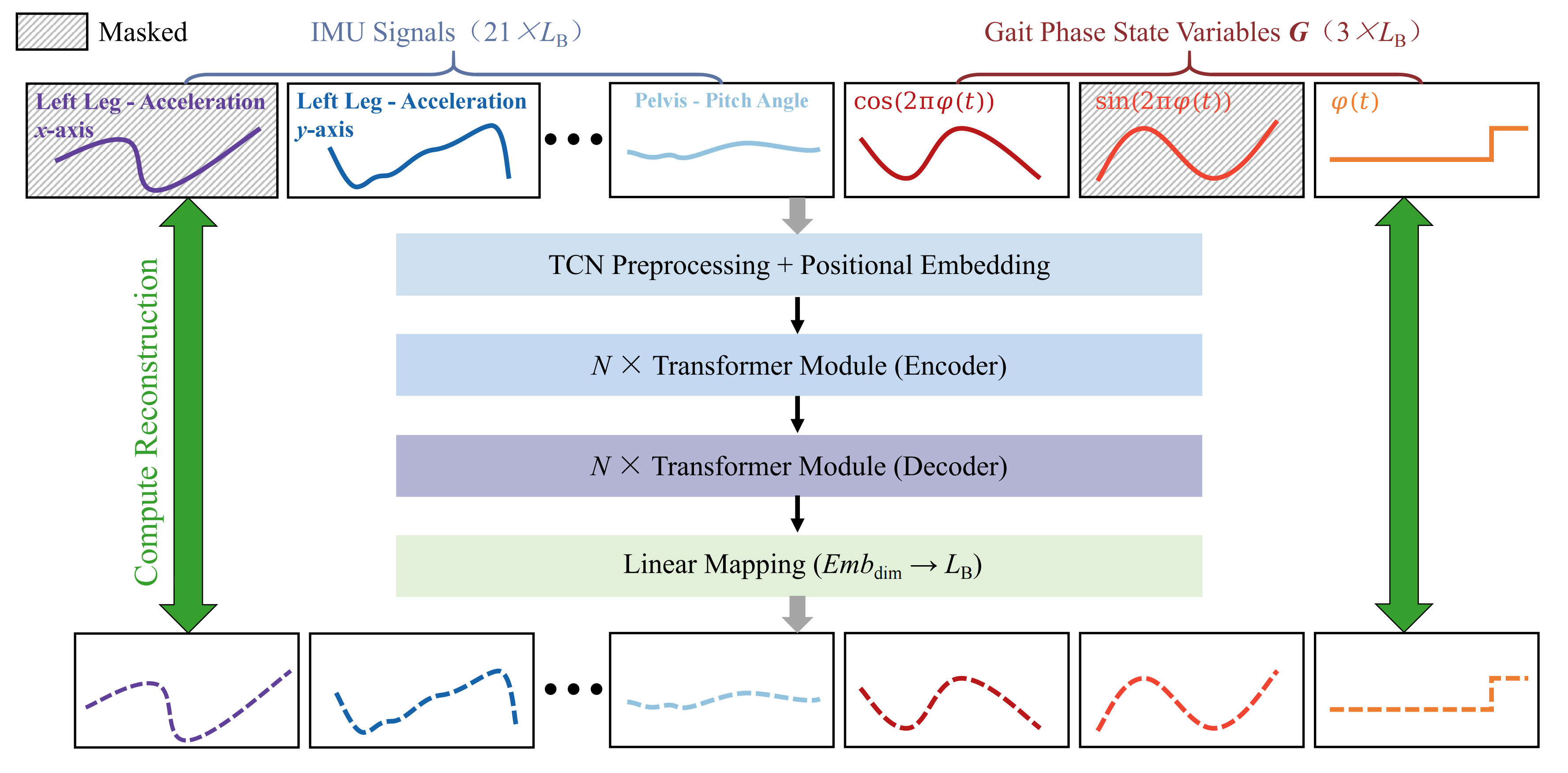}
\caption{Pre-training method based on implicit modeling of human locomotion. Several channels are randomly selected and masked by setting their values to zero. The masked time-series data are processed through the embedding layer, Transformer encoder layers, decoder layers, and a linear mapping layer to generate reconstructed time-series outputs. The reconstruction loss is computed only on the masked channels.}
\label{fig_3}
\end{figure*}

In theory, if an exact mathematical model of human locomotion were available, all other observable data could be inferred from a limited set of measurements. However, human locomotion is influenced by various factors, including terrain changes, voluntary motion intentions, and musculoskeletal system conditions, making it extremely complex and challenging to explicitly construct an accurate mathematical model. 
Mask Modeling (MM) provides an approach for implicit modeling of such complex systems. By randomly masking portions of the data and training the model to reconstruct the missing parts, MM enables learning of the underlying structure and feature relationships within the data without the need to explicitly model all contributing factors~\cite{ref31}. 
Based on this concept, this study proposes an implicit modeling framework for human locomotion using channel-wise MM, which is employed to pre-train the TCTST network. The detailed method is illustrated in Fig.~\ref{fig_3}.

\normalcolor

(1) \textit{Channel-wise Random Masking Strategy}: During human walking, the signals collected by IMUs across different channels can be considered as observations of the same underlying locomotion process. Likewise, the gait phase state vector, which describes the periodic evolution of this process over time, can also be viewed as an observation. It is reasonable to assume that when projected into a high-dimensional space, all channels encode equivalent information about the underlying motion. Based on this assumption, this study adopts a strategy of randomly masking a subset of channels. Specifically, for each training data sample, a masking ratio parameter \( M_\mathrm{R} \) is predefined. Then, \( C \times M_\mathrm{R} \) channels are randomly selected from the total \( C \) channels and their data are set to zero. The indices of the masked channels, along with their original unmasked data, are recorded for subsequent loss computation and evaluation.

(2) \textit{Encoder-Decoder Network Architecture}: The network architecture used for pre-training shares the same embedding and encoding layers as the proposed TCTST network. The key difference lies in the addition of an extra decoder composed of \( N \) Transformer layers following the \( N \) Transformer encoder layers. Finally, a simple linear mapping layer is applied to transform the feature representation from \( C \times Emb_\mathrm{dim} \) dimensions to a time-series output of size \( C \times L_\mathrm{B} \).

(3) \textit{Reconstruction Loss Computation}: In this study, the Mean Square Error (MSE) is used as the loss function to compute the reconstruction loss \( Loss_{\mathrm{recon}} \). The loss is calculated based on the indices of the masked channels recorded during the masking process. The detailed formulation is expressed as follows:

\begin{equation}
Loss_{\text{recon}} = 
\frac{1}{M} \sum_{i=1}^{M} \left(\frac{1}{L_\mathrm{B}} \sum_{n=1}^{L_\mathrm{B}} \left( c_{i,n} - \widetilde{c}_{i,n} \right)^2\right)
\end{equation}

where \( M \) denotes the number of masked channels, \( L_\mathrm{B} \) is the length of the look-back window, \( c_{i, n} \) represents the ground truth value at the \( n \)-th sampling point of the \( i \)-th masked channel, and \( \widetilde{c}_{i, n} \) denotes the corresponding reconstructed value.

\subsection{Pre-training and Fine-tuning Framework}
The proposed pre-training strategy is applied to train the model for 100 epochs, with an early stopping mechanism set to a patience of 10 epochs to prevent overfitting. During the pre-training process, two key hyperparameters are considered: the masking ratio \( M_\mathrm{R} \) and the stride \( s \) of the time-series sliding window. The masking ratio determines the proportion of input channels to be masked, while the stride controls the sampling frequency of the time-series data. The optimal combination of these two hyperparameters is explored in Section~\uppercase\expandafter{\romannumeral3}-D. During the inference phase, the three gait phase state channels are also padded with zeros and stacked along the channel direction with the IMU data to ensure that the input dimensions match the network structure.

\begin{figure*}[!t]
\centering
\includegraphics[width=0.85\textwidth]{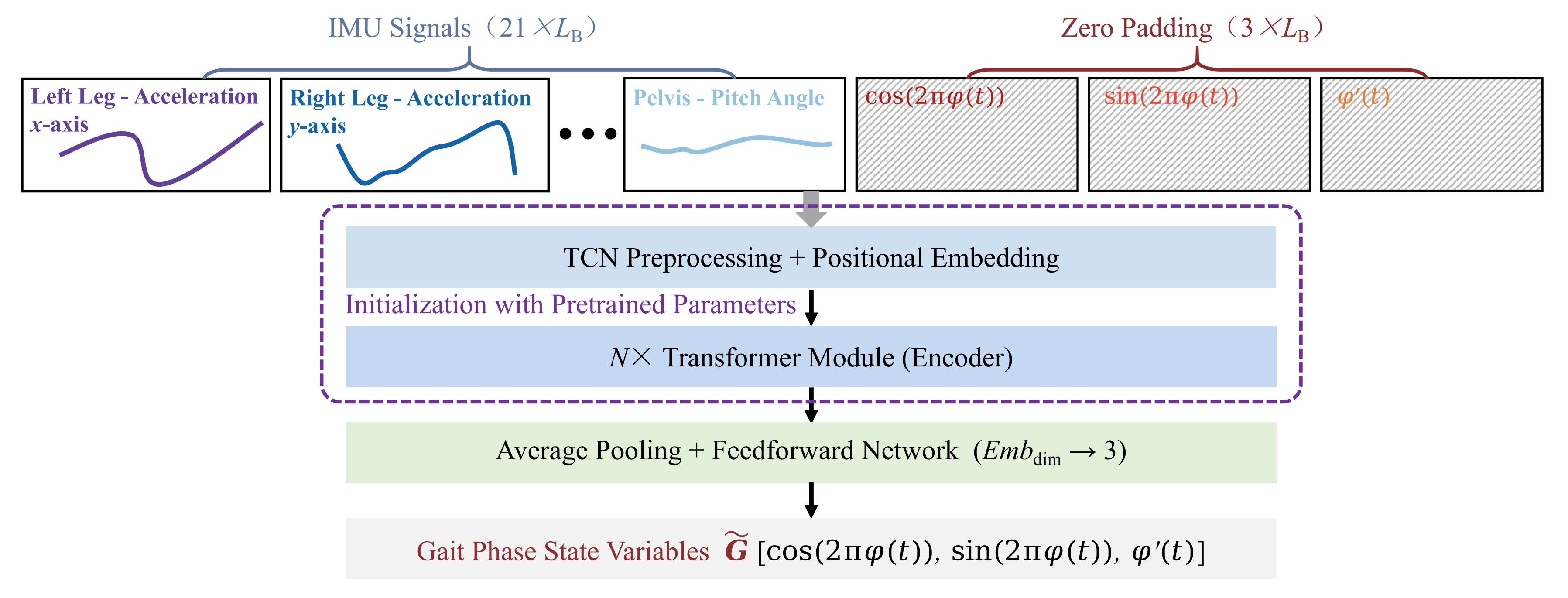}
\caption{Fine-tuning process after pre-training. A 3-channel zero vector is concatenated to the input for dimensional consistency. The embedding layer and Transformer encoder layers are initialized with the pre-trained parameters. The MSE between the estimated and ground-truth gait phase state vectors is computed for training.}
\label{fig_4}
\end{figure*}

After pre-training, the parameters learned from the pre-training stage are used to initialize the embedding layer and Transformer encoder layers of the model, allowing the network to fully leverage the feature representations learned during pre-training. The overall procedure is illustrated in Fig.~\ref{fig_4}. During the fine-tuning stage, the 21-channel IMU signals are concatenated with a 3-channel zero vector, resulting in a 24-channel input. The 3-channel zero vector serves as a placeholder input to maintain consistency with the input structure used in the pre-training stage, ensuring architectural alignment and reducing complexity associated with structural modifications. The fine-tuning process also runs for 100 epochs with an early stopping mechanism set to a patience of 10 epochs. The same MSE loss function is employed to compute the difference between the estimated and ground-truth gait phase state vectors during fine-tuning.

\normalcolor

\section{Experimental Setup and Results}
\subsection{Experimental Preparation}
This study adopts a Leave-One-Out Cross-Validation (LOO-CV) strategy for training and testing. Specifically, the dataset is divided into five folds based on subjects. In each fold, the training set contains data from seven subjects, the validation set contains data from one subject, and the test set contains data from two subjects, ensuring that each subject appears in the test set exactly once. 

The evaluation metrics include the Root Mean Square Error (RMSE), which measures the accuracy of gait phase estimation, and the Mean Absolute Error (MAE), which measures the accuracy of gait phase rate estimation. The performance of each algorithm is analyzed under stable terrain and terrain transitions. To assess the statistical significance of performance differences among algorithms, paired \textit{t}-tests are conducted using subject-wise results.

All network models are trained and evaluated on the same server running Ubuntu 22.04 with an NVIDIA RTX 4090 GPU. The code is implemented in Python 3.11.4 using the PyTorch 2.1.0 framework, and CUDA version 11.8 is employed for inference acceleration.

\subsection{IMU Time-Series Preprocessing Module}
The purpose of this experiment is to compare the performance of the TCN-based time-series preprocessing module with other commonly used preprocessing methods and to demonstrate the rationale behind choosing TCN as the preprocessing module. All other parameters in these models are kept identical across different preprocessing methods. The detailed experimental settings are as follows: the look-back window length is set to 100, the number of attention heads is 8, the number of Transformer layers is 8, and the embedding dimension is 384. The input consists of 21-channel IMU signals concatenated with a 3-channel zero vector.

The batch size during training is set to 1024, and the number of training epochs is 100. The learning rate is initialized to \(6 \times 10^{-4}\), and during the first 20 epochs, a cosine schedule is used to gradually increase the learning rate scaling factor from 0.2 to 1.0. Afterward, if the validation loss does not improve for 10 consecutive epochs, the learning rate is halved, with a minimum scaling factor of 0.1. An early stopping mechanism with a patience of 20 epochs is also employed.

Four different preprocessing methods are compared in this experiment: TCN, RNN, MLP, and Patch-based methods. The following describes the specific parameter configurations of each method. All preprocessing modules are designed in a channel-independent manner, mapping each channel of IMU time-series data from the input length of 100 to the embedding dimension \( Emb_\mathrm{dim} = 384 \).

\begin{itemize}
    \item[(1)] \textbf{MLP}: The first linear layer maps the input length of 100 to 192, followed by a ReLU activation function, and then a second linear layer maps 256 to the embedding dimension \( Emb_\mathrm{dim} = 384 \).
    \item[(2)] \textbf{RNN}: Two RNN layers are used, each with a hidden dimension of 192. The hidden state from the final time step is passed through a linear layer to obtain the embedding dimension \( Emb_\mathrm{dim} = 384 \).
    \item[(3)] \textbf{Patch-based}: The input sequence is divided into 10 non-overlapping patches of length 10. Each patch is projected to a 32-dimensional vector through a linear layer. All patches are concatenated to form a 320-dimensional intermediate vector, which is then passed through a Relu activation and a final linear layer to map to the embedding dimension \( Emb_\mathrm{dim} = 384 \).
\end{itemize}

The experimental results, summarized in Table~\ref{tab1}, show that the proposed TCN-based preprocessing module achieves the best performance. This may be attributed to its ability to effectively capture multi-scale temporal features along the time dimension. In contrast, both MLP and Patch are essentially composed of two linear transformations combined with activation functions to map input signals into feature space. Experimental results confirm that MLP and Patch-based methods exhibit similar performance. While both are capable of extracting preliminary feature representations, they lack the ability to capture temporal dependencies. Furthermore, using RNN for feature extraction prior to Transformer encoding may lead to an overly large and complex model, thereby increasing the difficulty of training and resulting in the worst performance among all methods.

\begin{table*}[!t]
\caption{Performance of Different Preprocessing Modules}
\label{tab1}
\centering
\renewcommand{\arraystretch}{1.3} 
\small 
\begin{tabular}{c cc cc}
\toprule
\multirow{2}{*}{Preprocessing Modules} & \multicolumn{2}{c}{Stable Terrain} & \multicolumn{2}{c}{Terrain Transition} \\
\cmidrule(lr){2-3} \cmidrule(lr){4-5}
 & RMSE (\%) & MAE (\%) & RMSE (\%) & MAE (\%) \\
\midrule
\textbf{TCN}   &  \textbf{2.947 $\pm$ 1.378} & \textbf{0.042 $\pm$ 0.022} & \textbf{3.364 $\pm$ 1.396} & \textbf{0.051 $\pm$ 0.024} \\
RNN   & 3.150 $\pm$ 1.034 & 0.047 $\pm$ 0.021 & 3.532 $\pm$ 1.215 & 0.055 $\pm$ 0.023 \\
MLP   & 3.094 $\pm$ 1.284 & 0.046 $\pm$ 0.024 & 3.512 $\pm$ 1.366 & 0.054 $\pm$ 0.024 \\
Patch & 3.085 $\pm$ 2.745 & 0.041 $\pm$ 0.023 & 3.396 $\pm$ 1.744 & 0.051 $\pm$ 0.024 \\
\bottomrule
\end{tabular}
\end{table*}
\normalcolor

\subsection{Hyperparameter Optimization of TCTST}
The purpose of this experiment is to identify the optimal set of network architecture hyperparameters for the TCTST model to serve as the foundation for subsequent experiments. The hyperparameters explored in this search include the embedding dimension \( Emb_\mathrm{dim}\), the number of attention heads \( n_{\mathrm{head}} \), and the number of Transformer encoder layers \( N_\mathrm{T} \). These three hyperparameters jointly determine the overall model size and representational capacity. 
The other experimental settings are as follows: the input consists of 21-channel IMU signals concatenated with a 3-channel zero vector; the batch size is set to 1024; the number of training epochs is 100; and the learning rate is initialized to \( 6 \times 10^{-4} \). 

A series of hyperparameter combinations are systematically evaluated, starting from smaller to larger values to gradually increase model complexity. Larger hyperparameter values generally indicate more complex network architectures with stronger representation capacity, but may also increase the risk of overfitting. The experimental results are summarized in Table~\ref{tab2}.

\begin{table*}[!t]
\caption{Experimental Results of Hyperparameter Search for TCTST Architecture}
\label{tab2}
\centering
\renewcommand{\arraystretch}{1.3} 
\small 
\begin{tabular}{ccc cc cc}
\toprule
\multirow{2}{*}{$Emb_\mathrm{dim}$} & \multirow{2}{*}{$n_\mathrm{head}$} & \multirow{2}{*}{$ N_\mathrm{T} $} 
& \multicolumn{2}{c}{Stable Terrain} 
& \multicolumn{2}{c}{Terrain Transition} \\
\cmidrule(lr){4-5} \cmidrule(lr){6-7}
 &  &  & RMSE (\%) & MAE (\%) & RMSE (\%) & MAE (\%) \\
\midrule
96  & 4  & 4  & 2.988 $\pm$ 1.390 & 0.042 $\pm$ 0.020 & 3.394 $\pm$ 1.367 & 0.052 $\pm$ 0.024 \\
192 & 6  & 6  & 2.972 $\pm$ 1.367 & 0.042 $\pm$ 0.023 & 3.376 $\pm$ 1.334 & 0.052 $\pm$ 0.024 \\
384 & 8  & 8  & 2.947 $\pm$ 1.378 & 0.042 $\pm$ 0.022 & \textbf{3.364 $\pm$ 1.396} & \textbf{0.051 $\pm$ 0.024} \\
384 & 12 & 12 & 3.020 $\pm$ 1.320 & 0.043 $\pm$ 0.022 & 3.415 $\pm$ 1.390 & \textbf{0.051 $\pm$ 0.024} \\
576 & 12 & 12 & 3.178 $\pm$ 1.462 & 0.045 $\pm$ 0.030 & 3.523 $\pm$ 1.397 & 0.054 $\pm$ 0.027 \\
768 & 12 & 12 & 3.095 $\pm$ 1.441 & 0.043 $\pm$ 0.024 & 3.483 $\pm$ 1.312 & 0.053 $\pm$ 0.023 \\
\bottomrule
\end{tabular}
\end{table*}
\normalcolor

\subsection{Pre-training Hyperparameter Search Experiments}
The purpose of this experiment is to identify the optimal hyperparameter configuration for pre-training. The hyperparameters explored include the masking ratio \( M_\mathrm{R} \) and the sliding window stride \( S_{\mathrm{Pre}} \) used during pre-training. The other experimental settings are as follows: the input consists of 21-channel IMU signals concatenated with a 3-channel zero vector; the batch size is set to 1024; the number of training epochs is 100; the learning rate is initialized to \( 6 \times 10^{-4} \); the embedding dimension \( Emb_\mathrm{dim}\) is 384; the number of attention heads \( n_{\mathrm{head}} \) is 8; and the number of Transformer encoder layers \( N_\mathrm{T} \) is 8.

\begin{figure}[!t]
\centering
\includegraphics[width=3.5in]{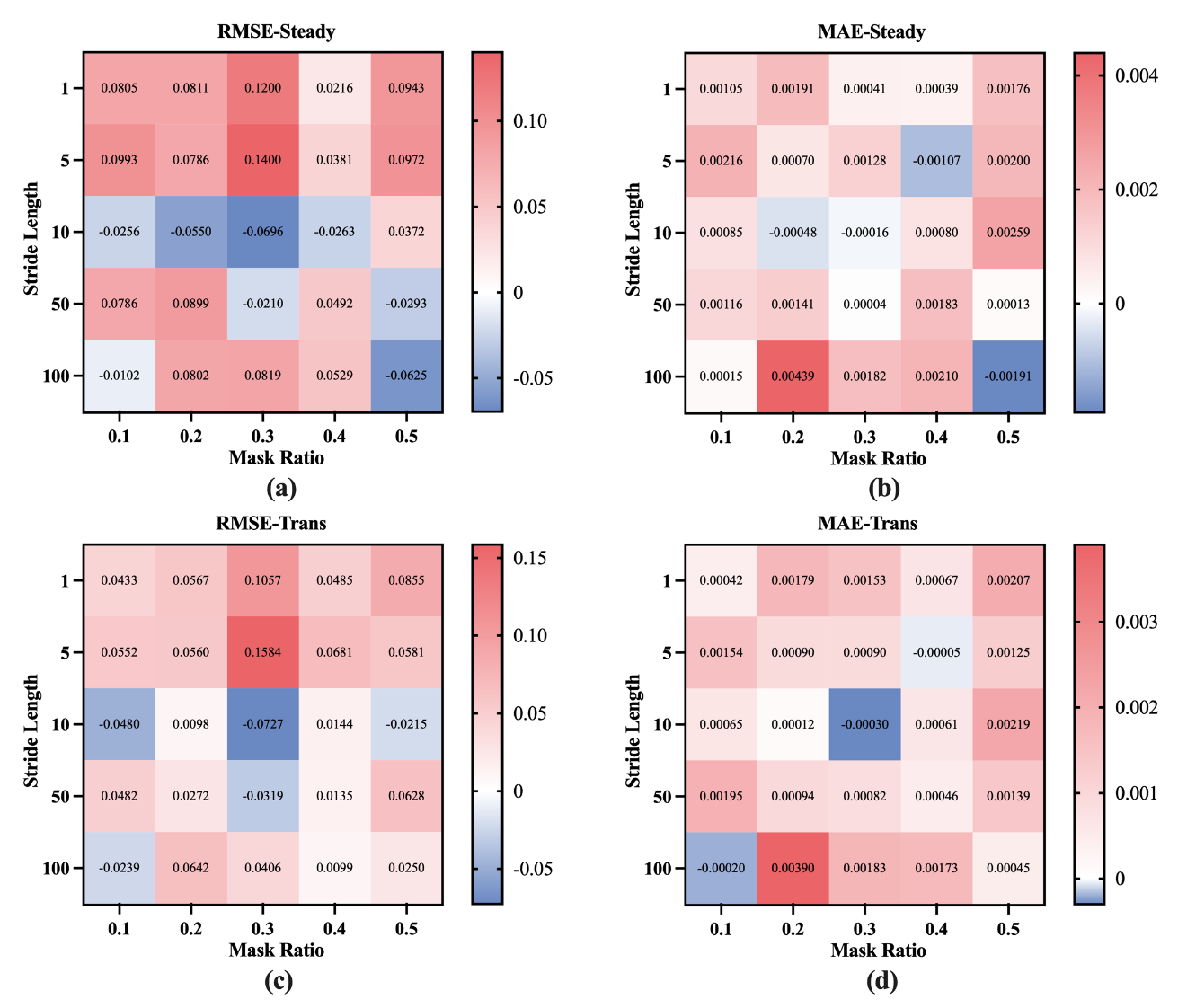}
\caption{Effect of pre-training hyperparameters on model performance. The values in each subfigure represent the difference between the network performance under each hyperparameter setting and the baseline performance without pre-training. The horizontal axis denotes the masking ratio, and the vertical axis represents the sliding window stride used during pre-training. (a) RMSE under stable terrain conditions. (b) MAE under stable terrain conditions. (c) RMSE under terrain transition conditions. (d) MAE under terrain transition conditions.}
\label{fig_5}
\end{figure}

As a baseline, we report the performance of a model trained from scratch without pre-training. The RMSE of gait phase estimation under stable terrain is \( 2.947 \pm 1.378\% \), and the MAE of gait phase rate estimation is \( 0.042 \pm 0.022\% \). Under terrain transitions, the RMSE of gait phase estimation is \( 3.364 \pm 1.396\% \), and the MAE of gait phase rate estimation is \( 0.051 \pm 0.024\% \). For each combination of pre-training hyperparameters, the performance improvement is calculated by subtracting the baseline value, and the results are visualized as a heatmap, as shown in Fig.~\ref{fig_5}.

The experimental results indicate that pre-training effectiveness is sensitive to hyperparameter selection. When the sliding window stride is too small, pre-training performance deteriorates, possibly because the high similarity between consecutive samples makes it difficult for the network to learn meaningful representations. When the masking ratio is too low, the network has too few patterns to learn; conversely, when the masking ratio is too high, the task becomes overly challenging, preventing the network from learning effective representations. Notably, when fixing \( S_{\mathrm{Pre}} = 10 \), we observe that the performance first improves and then degrades as \( M_\mathrm{R} \) increases, which aligns with theoretical expectations. Based on these findings, we select \( M_\mathrm{R} = 0.3 \) and \( S_{\mathrm{Pre}} = 10 \) as the optimal hyperparameter configuration for pre-training.

\normalcolor

\subsection{Ablation Study on the Pre-training Strategy}
The purpose of this experiment is to validate the effectiveness of the proposed pre-training design through four ablation studies. The specific experimental settings are as follows:

\begin{itemize}
    \item[(1)] \textbf{Decoder Ablation}: To investigate the role of the decoder layer in the pre-training process, we remove the decoder layer and examine its impact on model performance.
    \item[(2)] \textbf{Feature Level Reconstruction}: To explore whether reconstruction should be performed at the feature level or at the raw time-series level, we evaluate the reconstruction of feature vectors. Specifically, the features obtained from the embedding layer are treated as the original inputs, and the features output from the Transformer encoder and decoder are treated as the target for reconstruction.
    \item[(3)] \textbf{Non-channel-wise Masking}: To verify the effectiveness of the proposed channel-wise masking strategy, we conduct a comparative experiment using a two-dimensional random masking strategy applied over both channels and the look-back window.
    \item[(4)] \textbf{Gait Phase State Ablation}: To evaluate the hypothesis that the gait phase state vector can be regarded as one of the observable states of human locomotion, we test a variant where only the 21-channel IMU data are used for pre-training, and likewise only the 21-channel IMU data are used as input during fine-tuning.
\end{itemize}

Apart from the pre-training strategies under examination, all other parameters are kept consistent across experiments. Specifically, the masking ratio during pre-training is set to 0.3, the sliding window stride is 10, and the look-back window length is 100. The experimental results are summarized in Table~\ref{tab3}.

The experimental results demonstrate that each component of the proposed pre-training strategy contributes to the final model performance. Among all examined factors, the presence or absence of the decoder layer has the most significant impact, suggesting that introducing a decoder layer effectively enhances the network's ability to reconstruct time-series data.

\begin{table*}[!t]
\caption{Ablation Results of Pre-training Strategy}
\label{tab3}
\centering
\renewcommand{\arraystretch}{1.3} 
\small 
\begin{tabular}{c cc cc}
\toprule
\multirow{2}{*}{Method} & \multicolumn{2}{c}{Stable Terrain} & \multicolumn{2}{c}{Terrain Transition} \\
\cmidrule(lr){2-3} \cmidrule(lr){4-5}
 & RMSE (\%) & MAE (\%) & RMSE (\%) & MAE (\%) \\
\midrule
\textbf{Proposed} &  \textbf{2.877 $\pm$ 1.215} & \textbf{0.041 $\pm$ 0.019} & \textbf{3.291 $\pm$ 1.304} & \textbf{0.051 $\pm$ 0.022} \\
WDL   & 3.538 $\pm$ 1.398 & 0.046 $\pm$ 0.026 & 3.810 $\pm$ 1.216 & 0.054 $\pm$ 0.020 \\
FVR   & 2.956 $\pm$ 1.413 & 0.043 $\pm$ 0.024 & 3.358 $\pm$ 1.403 & 0.051 $\pm$ 0.024 \\
NCLM  & 3.018 $\pm$ 1.603 & 0.042 $\pm$ 0.021 & 3.386 $\pm$ 1.439 & 0.052 $\pm$ 0.023 \\
WPSV  & 2.977 $\pm$ 1.251 & 0.042 $\pm$ 0.019 & 3.386 $\pm$ 1.316 & 0.052 $\pm$ 0.022 \\
\bottomrule
\end{tabular}
\renewcommand{\arraystretch}{1.0} 

\vspace{2mm} 
\scriptsize Note: WDL = Without Decoding Layer, FVR = Feature Vector Reconstruction, NCLM = Non-Channel-Level Masking, WPSV = Without Phase State Variables.

\end{table*}
\normalcolor

\subsection{Comparison with Baseline Methods}
This study selects commonly used neural network algorithms in the field of gait phase estimation, including the bidirectional Long Short-Term Memory network (biLSTM) and Temporal Convolutional Network (TCN), and adapts the state-of-the-art time-series analysis method PatchTST~\cite{ref32} as baseline methods. In addition, the performance of models trained directly from randomly initialized parameters without any pre-training is also evaluated. All algorithms are tested under different look-back window lengths of 50, 100, 150, and 200 to assess their performance in estimating the gait phase state vector.

\begin{figure}[!t]
\centering
\includegraphics[width=3.5in]{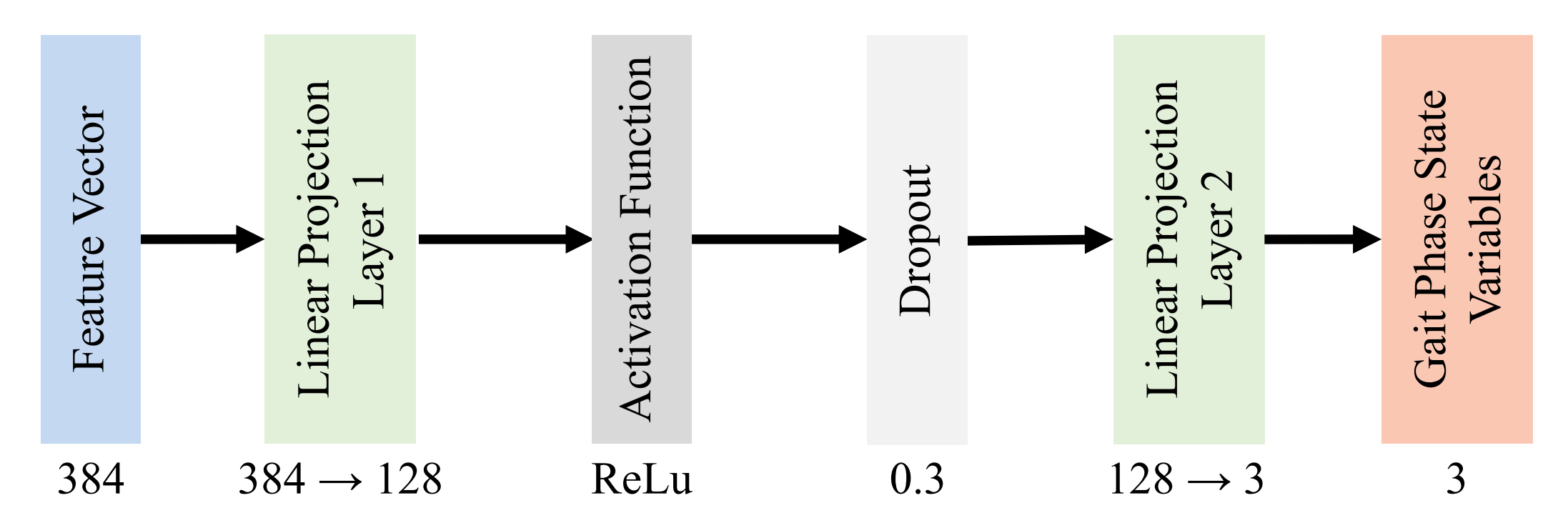}
\caption{Unified feedforward neural network adopted by each algorithm.}
\label{fig_6}
\end{figure}

To ensure fairness, all networks are trained with the same learning rate and share partially consistent design choices. Specifically:
\begin{itemize}
    \item[(1)] The dimensionality of hidden layers and feature vectors in all networks is set to 384.
    \item[(2)] A shared feedforward network (FFN) is used to map the feature vectors to the 3-dimensional gait phase state vector. As shown in Fig.~\ref{fig_6}, the FFN consists of a linear layer that reduces the dimensionality from 384 to 128, followed by a ReLU activation function, dropout regularization, and a final linear layer that reduces the dimensionality to 3.
\end{itemize}

\normalcolor

The specific designs of each baseline algorithm prior to the shared feedforward neural network (FFN) are described as follows:

\begin{itemize}
    \item[(1)] \textbf{biLSTM}: The input data are first processed by a four-layer bidirectional Long Short-Term Memory (biLSTM) network, with each layer containing 192 hidden units, to capture both forward and backward dependencies in the sequence. The hidden state at the final time step is extracted from the LSTM output as the feature vector. Since the network is bidirectional, the dimensionality of this feature vector is twice the number of hidden units, resulting in a 384-dimensional feature vector.
    
    \item[(2)] \textbf{TCN}: The input data are first passed through two one-dimensional convolutional layers with 32 and 64 channels, respectively. Each convolution has a kernel size of 3 and uses zero-padding to maintain the original sequence length. Each convolutional layer is followed by a ReLU activation function and a max-pooling layer with a pooling size of 2 to reduce the temporal dimension. After two rounds of pooling, the temporal length is reduced to one-fourth of the original length. The resulting feature maps are then flattened into a one-dimensional vector and projected to a 384-dimensional feature vector via a linear layer.
    
    \item[(3)] \textbf{PatchTST}: The input data are segmented into non-overlapping time patches of length 10 with a stride of 10, where each patch contains all channels. Each patch is flattened into a one-dimensional vector and projected to a 384-dimensional embedding through a linear layer, followed by the addition of positional embeddings. These embedded patches are then fed into a Transformer encoder with a depth of 8 layers. Each Transformer layer has 8 attention heads and a feedforward network (FFN) expansion ratio of 4. After Transformer encoding, the feature vectors of all patches are averaged to obtain a final 384-dimensional feature vector.
\end{itemize}

\begin{figure}[!t]
\centering
\includegraphics[width=3.5in]{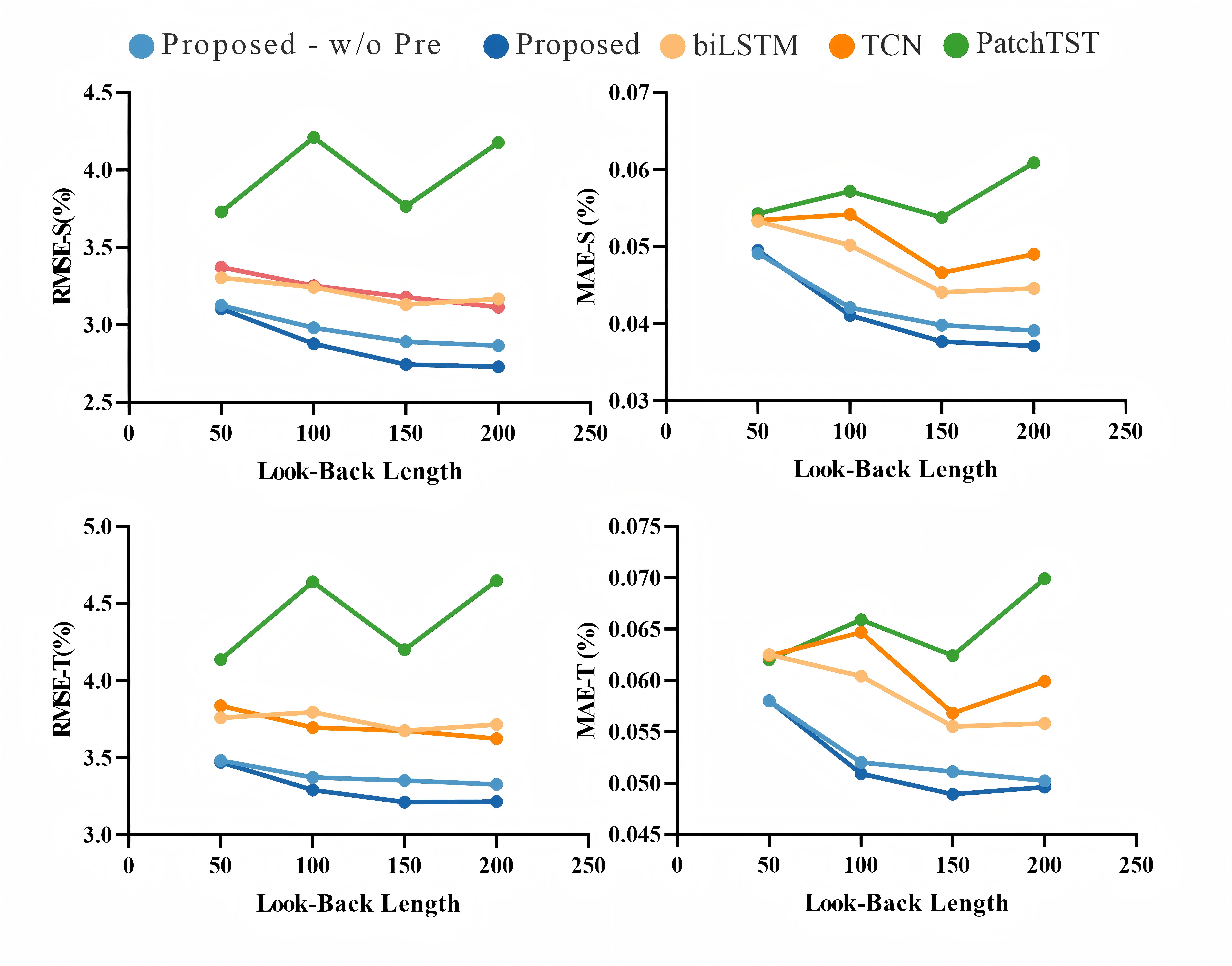}
\caption{Performance curves of different algorithms as a function of the look-back window length. From left to right and from top to bottom, the subfigures represent the RMSE of gait phase estimation and the MAE of phase rate estimation under stable terrain, followed by the RMSE of gait phase estimation and the MAE of phase rate estimation during terrain transition.}
\label{fig_7}
\end{figure}

The performance curves of each algorithm with respect to different look-back window lengths are plotted, as shown in Fig.~\ref{fig_7}
\normalcolor

These results demonstrate that the proposed TCTST with pre-training achieves the best overall performance. Furthermore, paired \textit{t}-tests are conducted to evaluate the statistical significance of the performance improvements over different baseline methods, as summarized in Table~\ref{tab4}.
\begin{table*}[!t]
\caption{Significance of Differences Between the Proposed Algorithm and Baseline Algorithms}
\label{tab4}
\centering
\renewcommand{\arraystretch}{1.3} 
\small 
\begin{tabular}{c c c c c c c c c c}
\toprule
\multirow{2}{*}{Algorithm} & \multicolumn{2}{c}{$L_\mathrm{B} = 50$} & \multicolumn{2}{c}{$L_\mathrm{B} = 100$} & \multicolumn{2}{c}{$L_\mathrm{B} = 150$} & \multicolumn{2}{c}{$L_\mathrm{B} = 200$} \\
\cmidrule(lr){2-3} \cmidrule(lr){4-5} \cmidrule(lr){6-7} \cmidrule(lr){8-9}
 & RMSE & MAE & RMSE & MAE & RMSE & MAE & RMSE & MAE \\
\midrule
biLSTM   & *** & *** & *** & *** & *** & *** & *** & *** \\
TCN      & *** & *** & *** & *** & *** & *** & *** & *** \\
PatchTST & *** & *** & *** & *** & *** & *** & *** & *** \\
Without Pre-training & ns & ns & * & ns & *** & *** & *** & *** \\
\bottomrule
\end{tabular}
\renewcommand{\arraystretch}{1.0} 

\vspace{2mm} 
\scriptsize Note: \textit{ns} = not significant; \textbf{*} indicates $p < 0.05$, \textbf{***} indicates $p < 0.001$.
\end{table*}
\normalcolor

The results of the statistical tests indicate that the proposed algorithm achieves significant performance improvements compared to existing baseline methods. Moreover, it is observed that the performance gains introduced by the proposed pre-training method are not statistically significant when using shorter look-back windows. However, as the look-back window length increases, the benefits of pre-training become more evident. This finding suggests that sufficiently long sequences (greater than 100 sampling points, approximately 1 second) are necessary to effectively perform implicit modeling of the human walking process.

\subsection{Comparison Across Subjects and Terrain Conditions}
\begin{figure}[!t]
\centering
\includegraphics[width=3.5in]{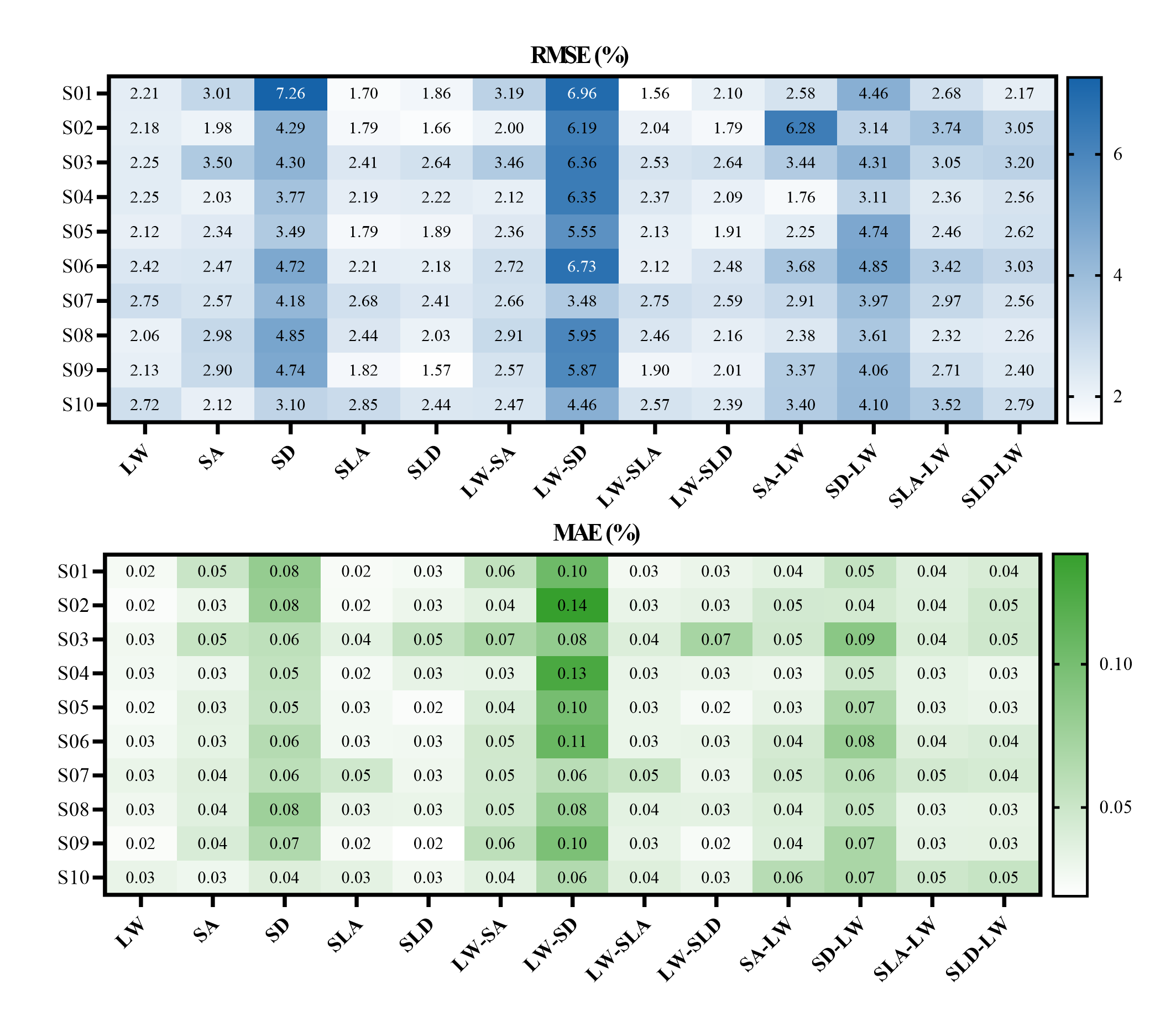}
\caption{Mean performance of the proposed algorithm across different subjects and terrains when $L_\mathrm{B} = 200$. The upper plot shows the RMSE of gait phase estimation, while the lower plot displays the MAE of phase rate estimation. Note: LW = Level Walking, SA = Stair Ascent, SD = Stair Descent, SLA = Slope Ascent, SLD = Slope Descent.}
\label{fig_8}
\end{figure}

To further analyze the performance differences of the proposed algorithm across different subjects and terrain conditions, heatmaps are plotted to visualize the gait phase estimation RMSE and gait phase rate estimation MAE for each subject under each terrain condition when \( L_\mathrm{B} = 200 \), as shown in Fig.~\ref{fig_8}.

Additionally, the average performance of the proposed algorithm across all subjects is summarized when \( L_\mathrm{B} = 200 \). Specifically, under stable terrain conditions, the RMSE of gait phase estimation is \( 2.729 \pm 1.071\% \), and the MAE of gait phase rate estimation is \( 0.037 \pm 0.016\% \); under terrain transition conditions, the RMSE of gait phase estimation is \( 3.215 \pm 1.303\% \), and the MAE of gait phase rate estimation is \( 0.050 \pm 0.023\% \).

\normalcolor

Analysis of the heatmap reveals that the trend of performance variation across terrain conditions is generally consistent among all subjects. The proposed algorithm performs better under stable terrain conditions compared to terrain transitions. Notably, the lowest performance is observed in stair descent-related terrains. Specifically, when walking on stable stair descent, the average RMSE is \( 4.17\% \) and the average MAE is \( 0.063\% \); during level-to-stair descent transitions, the average RMSE reaches \( 5.79\% \) and MAE is \( 0.096\% \); for stair descent to level ground transitions, the average RMSE is \( 3.68\% \) and MAE is \( 0.063\% \). Furthermore, the performance on stair ascent-related terrains is also lower compared to uphill and downhill terrains. This observation aligns with the fact that human gait exhibits greater variability and complexity when ascending or descending stairs.

\section{Real-World Validation on a Hip Exoskeleton}
To verify the practical effectiveness of the proposed gait phase estimation algorithm in exoskeleton assistance planning, we deployed the terrain-adaptive angle prediction network proposed in previous research~\cite{ref28} and the optimal model trained in the aforementioned study onto a system running Windows 11 with an RTX 4080 GPU laptop. One subject was recruited to wear a hip exoskeleton and a multimodal sensing system. The subject was informed of the procedures and potential risks before participation, and provided informed consent. The performance and computational cost of two terrain-adaptive trajectory planning schemes were tested: (1) selecting a predefined trajectory template based on estimated terrain, and (2) predicting the angular variation curve for the next step.

\subsection{Experimental Setup and Preparation}
\begin{figure}[!t]
\centering
\includegraphics[width=3.5in]{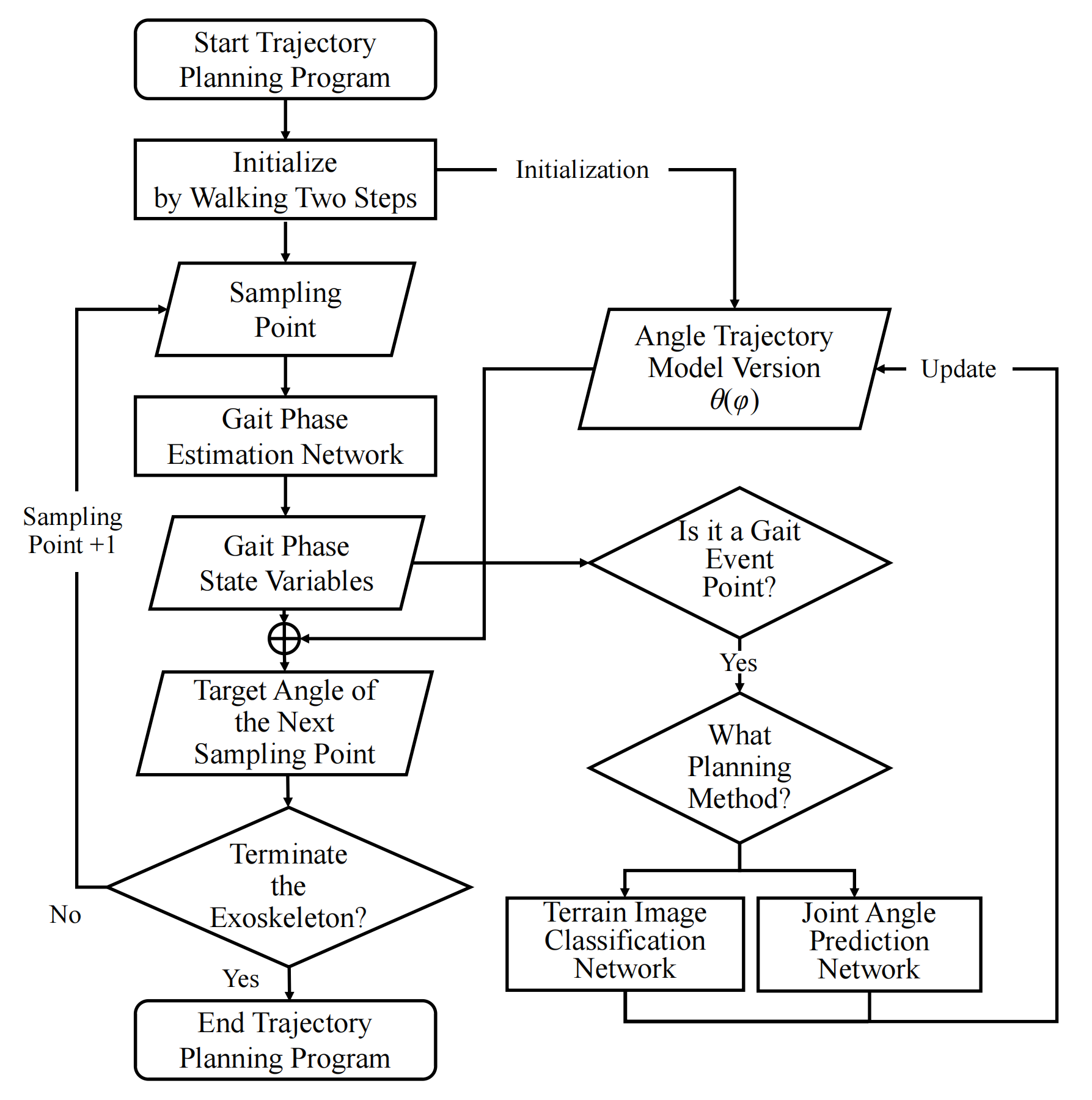}
\caption{Deep learning-based terrain-adaptive hip exoskeleton trajectory planning algorithm. The notation $\theta(\phi)$ in the figure represents the relationship between the planned joint angle and gait phase.}
\label{fig_9}
\end{figure}

The experiment consists of two parts: implementation of deep learning-based trajectory planning algorithms and wearable exoskeleton testing. The workflow of the trajectory planning algorithm is illustrated in Fig.~\ref{fig_9}, with a planning frequency synchronized to the IMU sampling rate at 100 Hz. The overall logic is as follows: first, the system initializes angle trajectory templates during the initial two steps. Then, during walking, for each sampling point, the system determines whether a gait event occurs. At each gait event, the angle trajectory template is updated. Based on the current gait phase state, the system predicts the thigh angle value at the next sampling point.

\normalcolor

\begin{itemize}
    \item[(1)] At each sampling point \( n \), the gait phase estimation network estimates the gait phase state vector \( \boldsymbol{G}(n) = (\cos \phi_n, \sin \phi_n, \phi'_n) \), which is then converted back to the gait phase state \( (\phi_n, \phi'_n) \) and recorded. A sampling point \( n \) is identified as a gait event point if \( \phi_n = 0\% \), or if \( \phi_{n-1} > 97\% \) and \( \phi_n < 3\% \).
    
    \item[(2)] At each gait event point, the system either (a) uses a terrain classification network to recognize the terrain and select a predefined angle trajectory template \( \theta(\phi) \) for updating, or (b) uses an angle trajectory prediction network to predict a step-specific trajectory \( \theta(\phi) \).
    
    \item[(3)] At each sampling point \( n \), the gait phase for the next sampling point is predicted as \( \phi_{n+1} = \phi_n + \phi'_n \). Based on this, the target control angle \( \theta(\phi_{n+1}) \) is generated.
\end{itemize}

\begin{figure}[!t]
\centering
\includegraphics[width=3.5in]{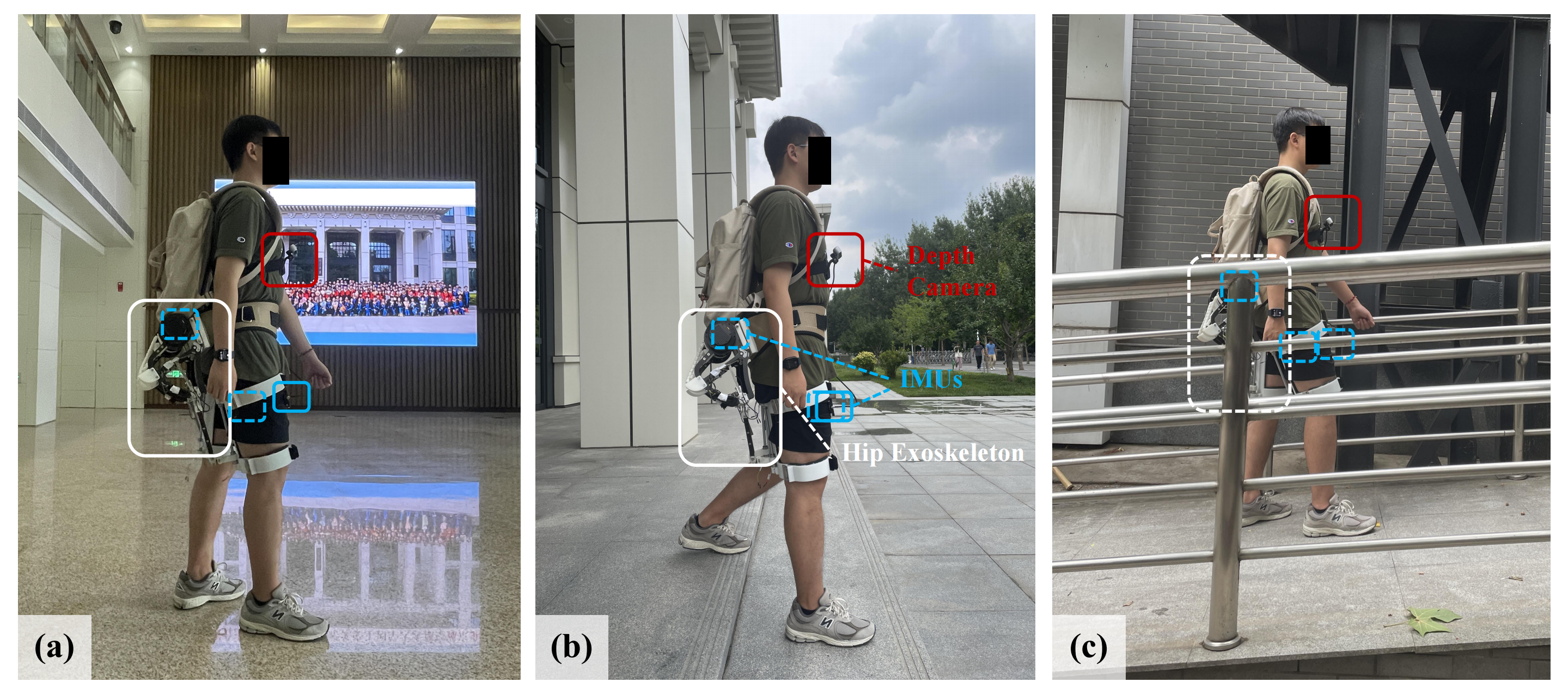}
\caption{Real-world validation of the hip exoskeleton system. From left to right, the images show testing on level ground (a), stair descent (b), and slope ascent (c).}
\label{fig_10}
\end{figure}

A walking experiment was conducted with one subject, covering five common types of terrain. During the experiment, the hip exoskeleton was worn but remained unpowered (passive mode). The system's main controller was a Windows-based laptop equipped with an NVIDIA RTX 4080 GPU. The experimental setup is illustrated in Fig.~\ref{fig_10}.

\normalcolor

\subsection{Computation Time Analysis of the Proposed Model}
The number of parameters (Para.), computational complexity (FLOPs), and inference latency (Latency) of each model are summarized in Table~\ref{tab5}. Considering that the real-time trajectory planning requires a frequency of 100 Hz---meaning that the prediction of the joint angle at the next sampling point must be completed within 10 ms after receiving the current data sample---the proposed gait phase estimation network achieves an inference latency of only 3.96 ms, which fully meets the real-time requirement.

\begin{table}[!t]
\caption{Parameter Size and Complexity of Each Model}
\label{tab5}
\centering
\renewcommand{\arraystretch}{1.3} 
\small 
\begin{tabular}{c c c c}
\toprule
Model & Para. (M) & Flops (G) & Latency (ms) \\
\midrule
Gait Phase Estimation   & 14.40  & 0.34  & 3.96  \\
Terrain Image Recognition & 5.74   & 0.67  & 10.09 \\
Joint Angle Prediction  & 46.06  & 9.51  & 11.84 \\
\bottomrule
\end{tabular}
\renewcommand{\arraystretch}{1.0} 
\end{table}
\normalcolor

Furthermore, both the terrain image classification network and the joint angle trajectory prediction network are used to update the trajectory template before each gait cycle. Although their inference latency slightly exceeds 10 ms, this would only introduce a negligible trajectory planning error of approximately 1\%.

\subsection{Trajectory Templates under Various Terrain Conditions}
\begin{figure}[!t]
\centering
\includegraphics[width=3.5in]{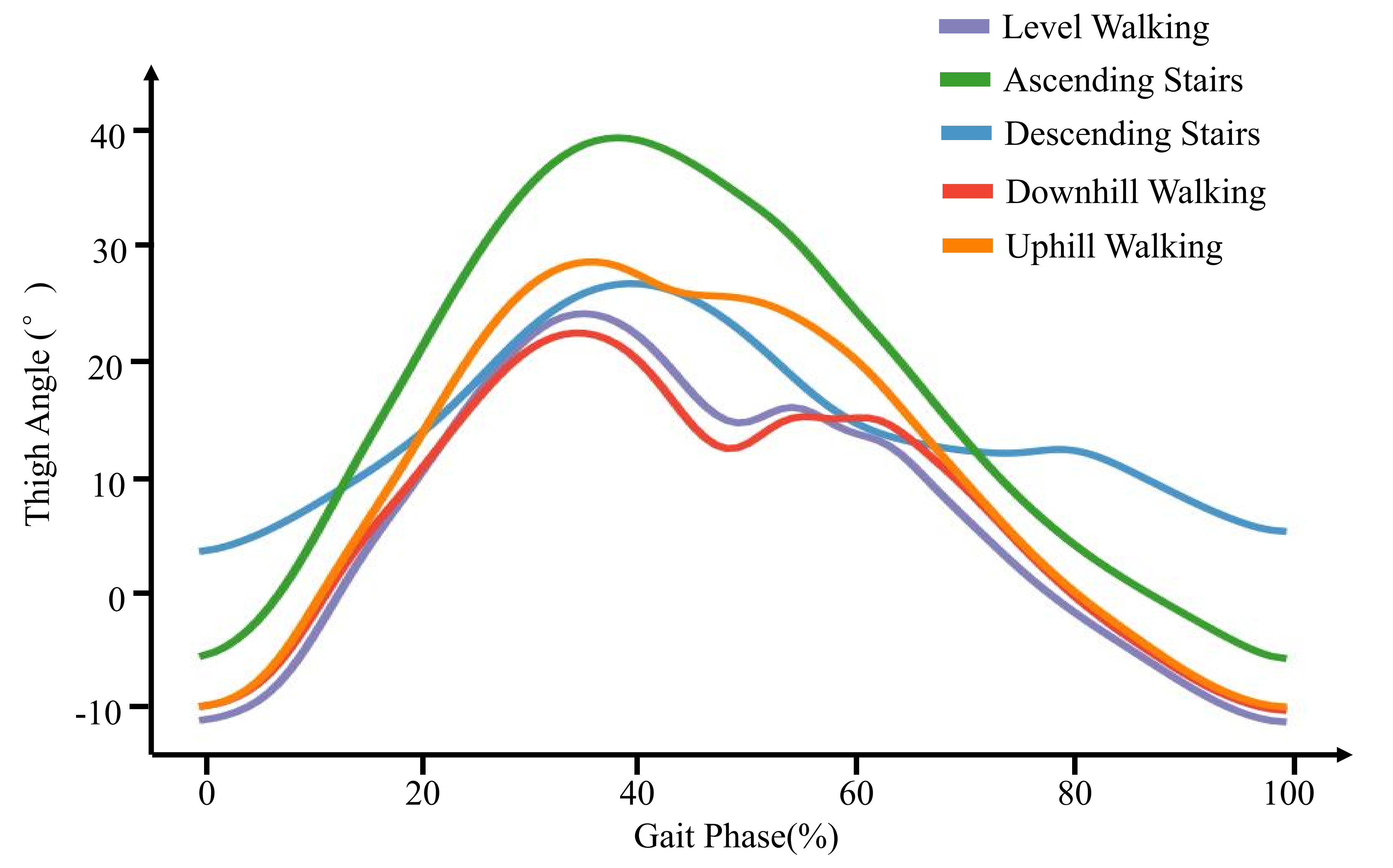}
\caption{Predefined trajectory templates corresponding to different terrains.}
\label{fig_11}
\end{figure}

Since the joint angle prediction network was not trained on data from the subject used in the hardware validation, the predefined trajectory templates for each terrain condition are set as the average trajectories obtained from the offline database collected in Section~\uppercase\expandafter{\romannumeral2}, as illustrated in Fig.~\ref{fig_11}. This ensures fairness in the evaluation process.

\normalcolor

\subsection{Results of Gait Phase Estimation}
\begin{figure}[!t]
\centering
\includegraphics[width=3.5in]{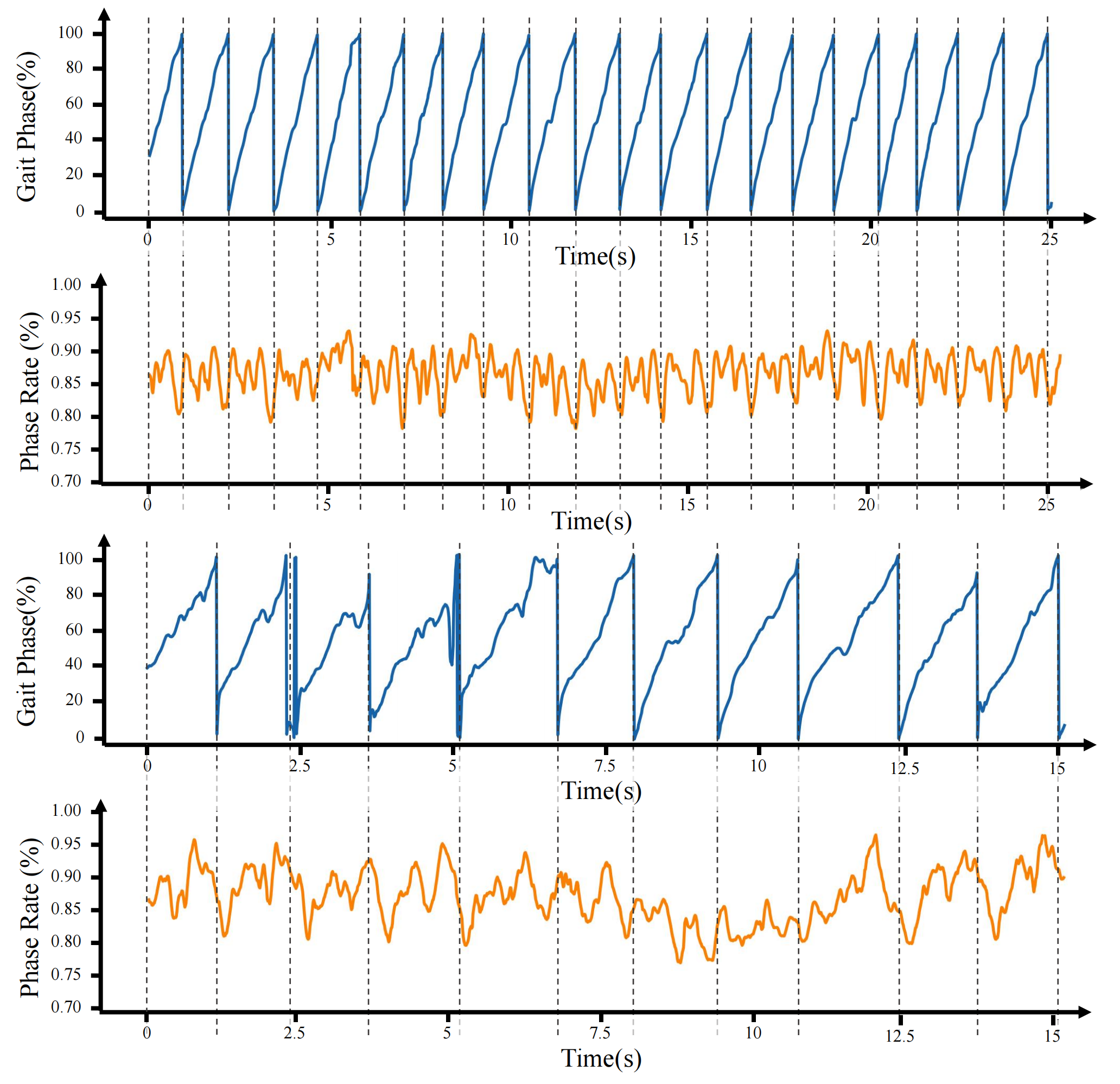}
\caption{Gait phase estimation results. From top to bottom, the subfigures represent the gait phase and phase rate under stable terrain, followed by the gait phase and phase rate during terrain transition.
}
\label{fig_12}
\end{figure}

Since ground-truth gait phase values are not available during hardware validation, this experiment follows the approach of Medrano \textit{et al.}~\cite{ref17}, only plots the estimated gait phase and gait phase rate obtained from the model, as shown in Fig.~\ref{fig_12}.

\normalcolor

The results indicate that the proposed algorithm achieves stable gait phase estimation under stable terrain conditions, with clearly distinguishable boundaries between consecutive gait cycles, enabling effective detection of gait events. Although the performance of gait phase estimation decreases during terrain transitions, the algorithm is still able to differentiate between distinct gait cycles.

\subsection{Comparison Between Planned and Measured Trajectories}
\begin{figure}[!t]
\centering
\includegraphics[width=3.5in]{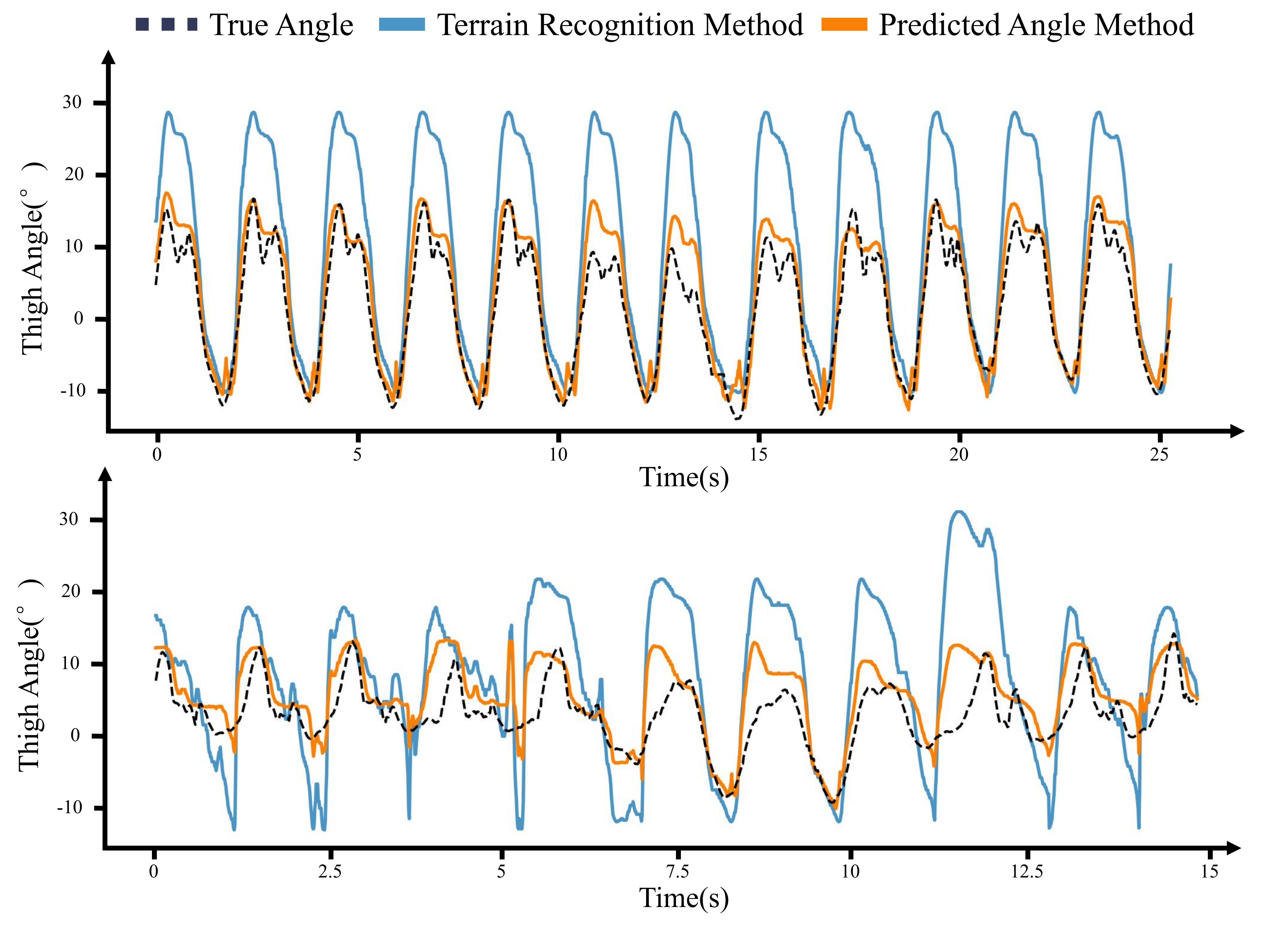}
\caption{Comparison between thigh trajectory planning and ground truth. The upper subfigure represents trajectory planning performance under stable terrain, while the lower subfigure represents performance during terrain transition.}
\label{fig_13}
\end{figure}

The objective of this study's trajectory planning is to enable the hip exoskeleton to follow natural human movements. Therefore, we evaluate the thigh angle tracking errors of two strategies: (1) terrain recognition with predefined trajectory template selection, and (2) step-wise joint angle prediction. The visualization focuses on the same time period as analyzed in Section~\uppercase\expandafter{\romannumeral4}-D, and the results are shown in Fig.~\ref{fig_13}.

\normalcolor

\begin{figure}[!t]
\centering
\includegraphics[width=3.5in]{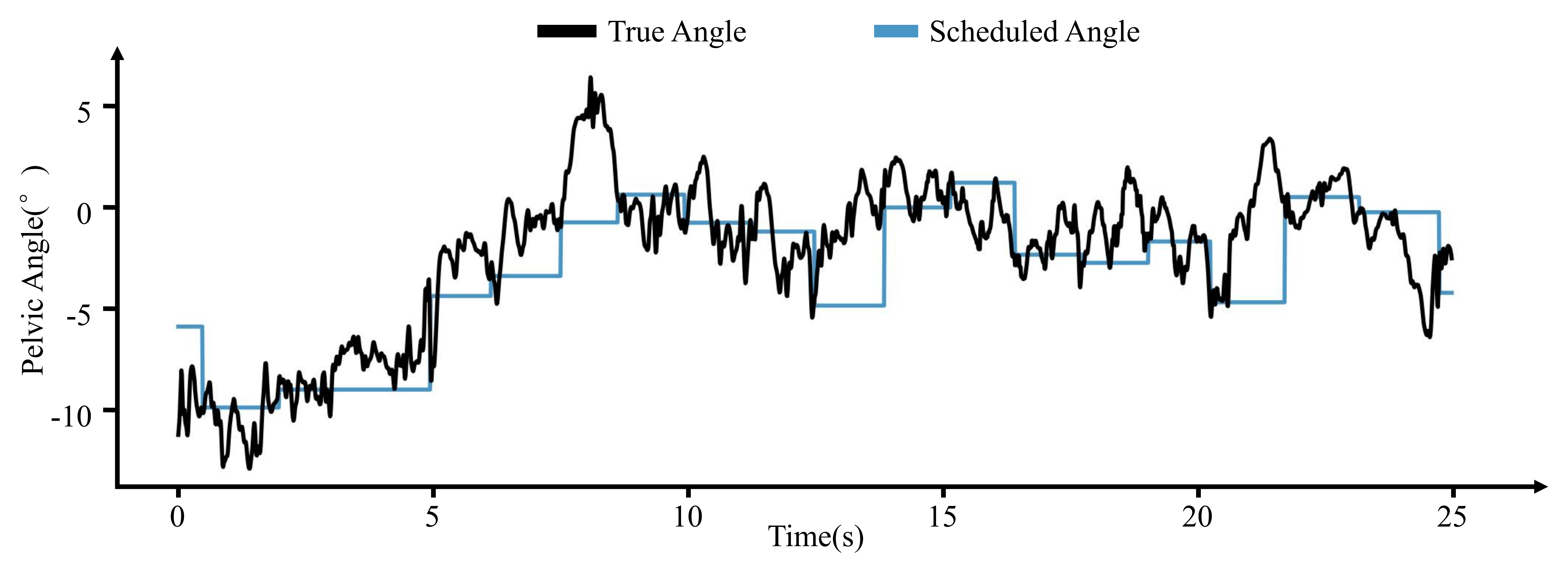}
\caption{Comparison between pelvis trajectory planning and ground truth.}
\label{fig_14}
\end{figure}

The quantitative analysis results show that the terrain recognition strategy yields an RMSE of \( 10.343^\circ \) and a Pearson correlation coefficient (PCC) of 0.887, while the joint angle prediction strategy achieves an RMSE of \( 4.187^\circ \) and a PCC of 0.922. These results indicate that the terrain recognition method performs poorly due to a large discrepancy between the subject's relatively small step length and the predefined trajectory templates. In contrast, the joint angle prediction method effectively adapts to variations in step length, resulting in predicted angles that closely match the actual values, and also performs well during terrain transitions. The trajectory planning result that treats the pelvic angle as a constant is illustrated in Fig.~\ref{fig_14}.

\normalcolor

The experimental results show that treating the pelvic angle as a constant can achieve relatively good performance. The ground-truth pelvic angle exhibits random fluctuations around the planned value, with an RMSE of \( 5.231^\circ \) and a Pearson correlation coefficient (PCC) of 0.466. 

It is noted that large discrepancies between the ground-truth and planned pelvic angles occur around 8 s and 20 s. Observing the trend, these discrepancies are characterized by a significant forward tilt of the subject’s upper body, which likely corresponds to transitions to upward terrains such as stair ascent and uphill walking. 

Since this study considers both thigh and pelvic angles separately, the effect of terrain on hip joint trajectories is primarily captured and tracked through the thigh angle predicted by the model. Therefore, these deviations in pelvic angle have a limited impact on the final planned trajectories.

\section{Discussion and Conclusion}
The gait phase estimation network designed in this study is a time-series analysis network. Considering the recent focus in time-series analysis on effective pre-training strategies, this study proposes an implicit modeling approach for human motion. First, a neural network architecture was designed, employing a Temporal Convolutional Network (TCN) for feature extraction and a Transformer for inter-channel information fusion. The proposed network structure outperforms most baseline methods. Furthermore, this study explores pre-training techniques for the network and introduces a masked reconstruction approach to achieve implicit modeling of human walking motion. Experimental results demonstrate the effectiveness of the proposed strategy, with the designed algorithm significantly outperforming baseline methods. When the look-back window length is set to 200, the proposed algorithm achieves a gait phase estimation root mean square error (RMSE) of $2.729 \pm 1.071\%$ and the phase rate estimation mean absolute error (MAE) of $0.037 \pm 0.016\%$ under stable terrain conditions. During terrain transitions, the gait phase estimation RMSE is $3.215 \pm 1.303\%$, and the phase rate estimation MAE is $0.050 \pm 0.023\%$.

Furthermore, a hip exoskeleton trajectory planning program was developed based on the three aforementioned neural networks, including the gait phase estimation network, terrain recognition network, and joint angle prediction network, and real-world validation was conducted with a wearable hip exoskeleton. Under stable terrain conditions, the system effectively estimated the gait phase, and the planned angular trajectory closely matched the natural human motion trajectory. Although during terrain transitions, performance exhibited some degree of degradation. The results indicate that the joint angle prediction method achieves superior performance compared to the predefined trajectory selection method based on terrain classification, with the RMSE and Pearson correlation coefficient (PCC) of the terrain classification-based method being $10.343^\circ$ and $0.887$, respectively, while the joint angle prediction method achieves an RMSE of $4.187^\circ$ and a PCC of $0.922$.

Future research directions may focus on the following aspects. First, in terms of the gait phase estimation algorithm, a current limitation is the lack of terrain-awareness. Experimental results indicate a noticeable performance drop when encountering stair ascent/descent terrains. Effectively integrating first-person vision information to anticipate upcoming terrain changes could enhance gait phase estimation accuracy across different terrains. Second, in terms of exoskeleton planning and control systems, designing an effective controller that reduces users’ metabolic cost, improves locomotion performance, and enhances exoskeleton usability across diverse terrains remains a critical challenge.

\end{document}